# EdgeLDR: Quaternion Low-Displacement Rank Neural Networks for Edge-Efficient Deep Learning

**Vladimir Frants[1], Sos Agaian[2], Life Fellow, IEEE and Karen Panetta[1], Member, IEEE**
[1]Department of Electrical and Computer Engineering, Tufts University, Medford, MA 02155, USA
[2]Department of Computer Science, College of Staten Island, The City University of New York, Staten Island, NY 10314, USA

Corresponding author: Vladimir Frants (e-mail: vladimir.frants@tufts.edu).

**ABSTRACT** Deploying deep neural networks on edge devices is often limited by the memory traffic and compute cost of dense linear operators. While quaternion neural networks improve parameter efficiency by coupling multiple channels through Hamilton products, they typically retain unstructured dense weights; conversely, structured matrices enable fast computation but are usually applied in the real domain. This paper introduces EdgeLDR, a practical framework for quaternion block-circulant linear and convolutional layers that combines quaternion channel mixing with block-circulant parameter structure and enables FFT-based evaluation through the complex adjoint representation. We present reference implementations of EdgeLDR layers and compare FFT-based computation against a naive spatial-domain realization of quaternion circulant products. FFT evaluation yields large empirical speedups over the naive implementation and keeps latency stable as block size increases, making larger compression factors computationally viable. We further integrate EdgeLDR layers into compact CNN and Transformer backbones and evaluate accuracy–compression trade-offs on 32×32 RGB classification (CIFAR-10/100, SVHN) and hyperspectral image classification (Houston 2013, Pavia University), reporting parameter counts and CPU/GPU latency. The results show that EdgeLDR layers provide significant compression with competitive accuracy.

## I. INTRODUCTION

Deep neural networks now deliver state-of-the-art performance in vision, speech, language, and scientific machine learning, but their expanding parameter counts and compute demands make them ill-suited for deployment on edge devices [1]–[4]. Mobile phones, wearable devices, IoT sensors, AR/VR headsets, and embedded controllers typically offer only a few megabytes of on-chip memory, limited off-chip bandwidth, and tight power budgets, often without access to powerful GPUs or stable high-bandwidth connectivity [2]–[6]. In such settings, dense, unstructured layers whose storage and compute scale as $O(n^2)$ result in latency and energy overheads that are prohibitive for always-on and real-time workloads [6]–[8].

These constraints are particularly acute for edge-relevant tasks that must continuously process multi-channel sensor streams: keyword spotting and on-device speech recognition, on-device vision for gesture recognition and object detection, environmental and audio event recognition, monitoring of industrial and home-automation sensors, and low-latency tracking in AR/VR pipelines [5], [9]. Fetching the parameters of large dense layers can dominate both latency and energy, and the highly regular, massively parallel compute patterns assumed by server-class accelerators are poorly suited to heterogeneous, resource-limited edge hardware [2], [3], [7], [8]. This inspires the creation of layers that are compact, leverage fast transforms, and match the structure of typical multi-channel sensor streams.

Structured matrices offer a principled way to couple compression with fast algorithms. Classic examples such as Toeplitz, Hankel, circulant, block-circulant, and other Toeplitz-like families admit $O(rn)$ parameterizations and support $O(rn\log n)$ matrix–vector products via FFT-based transforms for $n \times n$ matrices with displacement rank $r$ [10]–[12]. Other families, including Vandermonde and Cauchy matrices, also admit sub-quadratic fast algorithms based on polynomial and rational transform techniques [13], [14].

Building on this line of work, Sindhwani et al. and Zhao et al. showed that replacing dense weights with Low-Displacement Rank (LDR) matrices yields small-footprint networks with fast inference, while preserving universal approximation properties [15], [16]. Block-circulant architectures such as CirCNN and its extensions exploit real block-circulant matrices and FFTs to compress and accelerate fully-connected and convolutional layers in practice [17]–[20].

Real-world applications often benefit from inductive biases introduced by architectural choices: periodic video analysis for human action recognition and remote photoplethysmography can exploit quasi-periodic temporal structure [21], [22], rain removal networks leverage spatially repeated streak patterns [23], [24], and panoramic or omnidirectional vision models benefit from circulant padding or circulant convolutions to handle wrap-around image domains [25], [26]. Beyond temporal and spatial structure, many multi-channel sensing modalities exhibit strong inter-channel correlations - across color channels, spectral bands, or multi-sensor streams, which motivates architectures that explicitly group and jointly transform related channels rather than treating them as independent scalars.

Studies show that Quaternions Deep Neural Networks (QDNNs) have emerged as a powerful machine learning technique in numerous artificial intelligence applications, often demonstrating superior performance and parameter efficiency compared to their real-valued counterparts [27]–[35]. However, the standard QDNNs still employ large, dense weight matrices. Their large model sizes, despite offering better performance, make them computation- and memory-intensive, thus limiting their practical deployment on resource-constrained edge devices.

A transformative solution requires combining, for the first time, a structural compression technique, such as a block-circulant structure, with the inherent multi-channel efficiency of a QDNN system to create a unified framework for Edge-Efficient Deep Learning using quaternion algebra. This integration must also effectively leverage Fast Fourier Transform (FFT)-based acceleration within the QDNN system, necessitating a reliable mapping from quaternion-valued to complex-valued matrices for efficient FFT processing.

At the same time, recent work in quaternion linear algebra has developed quaternion Fourier transforms and shown that quaternion circulant and block-circulant matrices are block-diagonal [36]–[39]. These theoretical results strongly suggest the feasibility of structured quaternion operators. However, they have not yet been translated into trainable layers with efficient forward and backward passes, nor have they been rigorously connected to the established LDR theory.

To address these critical gaps, we have developed a general quaternion LDR framework for deep linear layers, focusing on the construction of the Block-Circulant Quaternion Fully-Connected (BC-QFC) and Convolutional (BC-QConv) layers. Our proposed layers operate directly on quaternion-valued activations, exploit multi-channel structure via the Hamilton product, and impose the LDR structure to enable powerful FFT-based acceleration. When integrated into compact CNN-style or Transformer-style architectures for edge-relevant tasks, we refer to them collectively as EdgeLDR layers. Our main contributions are summarized as follows:

1. We present the Quaternion Low-Displacement-Rank (LDR) Framework by extending the traditional displacement-operator perspective to quaternion-valued weight matrices. This involves defining the quaternion LDR structure via real displacement operators and introducing an efficient quaternion-to-complex mapping through the complex adjoint representation, which is crucial for enabling complex-domain computation.
2. We develop and analyze the highly accelerated structured quaternion layers: the Block-Circulant Quaternion Fully-Connected and Block-Circulant Quaternion Convolutional layers. We derive the complete, efficient FFT-based methods for their forward and backward passes, achieving a state-of-the-art computational complexity of $\mathcal{O}(n \log n)$.
3. We systematically incorporate the EdgeLDR layers into compact deep learning architectures and perform extensive evaluation. We assess their performance on diverse edge-relevant tasks, including RGB image classification (CIFAR-10/100, SVHN), hyperspectral image classification (Houston 2013 and Pavia University, comparing them comprehensively against standard dense, low-rank, pruning/quantization, and real block-circulant baselines.

The rest of the paper is organized as follows. Section II reviews edge-efficient neural networks, structured matrices, low-displacement-rank theory, and quaternion neural networks. Section III presents the quaternion LDR framework, including quaternion algebra, Fourier transforms, and block-circulant matrices, and derives the FFT-based forward and backward passes. Section IV reports experimental results on RGB and hyperspectral image classification. Section V discusses the broader implications and design considerations. Section VI concludes the paper.

## II. RELATED WORK

Model compression and acceleration techniques address these constraints through several complementary approaches. Low-rank factorization methods approximate weight matrices by lower-rank decompositions to reduce the number of parameters [40]. Pruning techniques directly remove redundant weights or entire channels/filters, effectively shrinking the model by trimming less meaningful connections [41], [42]. Quantization of network parameters/activations to low precision, for example, 8-bit integers or binary bits, significantly lowers memory usage and inference computation, often with minimal loss in accuracy [43]–[45]. Another major direction is knowledge distillation, where a large "teacher" model's knowledge is used to train a smaller "student" model, transferring the performance of the big model to a compact network [46]. Collectively, these methods can significantly compress deep models and speed up inference while maintaining high accuracy.

In parallel, researchers have designed efficient neural network architectures tailored for embedded and mobile environments. Instead of compressing a pre-existing model, these architectures build efficiency into the topology from the outset. Notable examples include the MobileNet family, which employs depthwise separable layers to decrease computations significantly, and ShuffleNet, which utilizes grouped convolutions and channel shuffling for improved efficiency [47], [48]. These models achieve competitive accuracy on vision tasks with far fewer parameters and FLOPs, making them well-suited for real-time inference on resource-constrained devices.

Complementary to these strategies, structured transforms replace dense weight matrices with algebraically structured matrices that admit fast algorithms [15], [49], [50]. A comprehensive survey can be found in [49]. Asriani et al. adapted the block-circulant structure and DCT/DST transforms to Transformers [18]. Qin et al. combined block-circulant weights with power-of-two quantization [19]. Zhou et al. [20] applied block-circulant matrices to accelerate GNNs. More recently, Samudre et al. used structured matrices, including LDR classes, to construct efficient approximately equivariant networks [51].

While the above methods operate on real or complex-valued weights, hypercomplex neural networks extend real-valued architectures by letting activations and weights live in algebras such as complex, quaternion, or octonions [27], [30], [33]–[35], [52], [53]. This enhances the modeling of inter-channel dependencies while using fewer degrees of freedom compared to unconstrained real layers.

At the same time, advances in quaternion linear algebra and Fourier analysis have clarified the spectral structure of circulant, block-circulant, and Toeplitz operators in the quaternion domain, paving the way for structured quaternion transforms and fast algorithms. Pan and Ng studied quaternion circulant matrices and showed that they cannot, in general, be simultaneously diagonalized by a single quaternion Fourier matrix in the same way the DFT diagonalizes complex circulant matrices [36]. Instead, they admit block-diagonal forms with 1×1 and 2×2 quaternion blocks. Zheng and Ni extended these block-diagonalization results to quaternion block-circulant matrices and used them to accelerate T-products of quaternion tensors, with applications to color video processing [37]. Sfikas et al. analyzed the matrix form of the quaternion Fourier transform and quaternion convolution, introducing a complex 2×2 block (complex adjoint) representation that realizes quaternion Fourier operations via pairs of complex FFTs [38]. Lin and Ng studied Hermitian quaternion Toeplitz matrices generated by quaternion-valued functions, providing spectral characterizations and linking quaternion Toeplitz structure to displacement-rank concepts [39].

## III. METHOD

We now present the quaternion LDR framework underlying our EdgeLDR layers. We first review quaternion algebra and its complex adjoint representation, then introduce quaternion Fourier transforms, structured quaternion matrices, and the resulting block-circulant quaternion layers with FFT-based algorithms.

Many classical structured matrices can be understood via the low-displacement-rank (LDR) framework [15], [16], [49], [50]. Given fixed structural matrices $A \in \mathbb{R}^{m\times m}$ and $B \in \mathbb{R}^{n\times n}$, the Sylvester-type displacement operator

$$\mathcal{L}_{A,B}(M) = AM - MB, \qquad M \in \mathbb{R}^{m\times n} \quad (1)$$

measures structure through the rank of $\mathcal{L}_{A,B}(M)$. A matrix $M$ is said to have displacement rank $\rho$ if $\mathcal{L}_{A,B}(M)$ has rank $\rho$. Toeplitz and Hankel matrices have displacement rank at most two for appropriate shifts $(A,B)$; Vandermonde and Cauchy matrices also admit very low displacement rank [15], [50]. Representing $M$ via generators $G \in \mathbb{R}^{m\times\rho}$, $H \in \mathbb{R}^{n\times\rho}$ such that $\mathcal{L}_{A,B}(M) = GH^\top$ yields $O\big(\rho(m+n)\big)$ parameters and enables matrix–vector products in $O(\rho n \log n)$ time using FFT-like transforms [15], [16].

### A. QUATERNION ALGEBRA AND COMPLEX ADJOINT REPRESENTATION

We work in the quaternion division algebra $\mathbb{H}$, viewed as a four-dimensional real associative algebra with canonical basis $\{1,\mathbf{i},\mathbf{j},\mathbf{k}\}$. The Hamilton product on $\mathbb{H}$ is determined by $\mathbf{i}^2 = \mathbf{j}^2 = \mathbf{k}^2 = \mathbf{ijk} = -1$, so that in particular $\mathbf{ij} = \mathbf{k}$, $\mathbf{jk} = \mathbf{i}$, $\mathbf{ki} = \mathbf{j}$, $\mathbf{ji} = -\mathbf{k}$, $\mathbf{kj} = -\mathbf{i}$, $\mathbf{ik} = -\mathbf{j}$, reflecting the noncommutativity of the product. Every quaternion $q \in \mathbb{H}$ can be written uniquely as $q = a_0 + a_1\mathbf{i} + a_2\mathbf{j} + a_3\mathbf{k}$ with $a_\ell \in \mathbb{R}$; its conjugate is $\bar{q} = a_0 - a_1\mathbf{i} - a_2\mathbf{j} - a_3\mathbf{k}$, and the norm is $\|q\| = \sqrt{a_0^2 + a_1^2 + a_2^2 + a_3^2}$. If $q = a_0 + a_1\mathbf{i} + a_2\mathbf{j} + a_3\mathbf{k}, p = b_0 + b_1\mathbf{i} + b_2\mathbf{j} + b_3\mathbf{k}$, then their Hamilton product $qp$ is given by:

$$\begin{aligned} qp \;&= (a_0b_0 - a_1b_1 - a_2b_2 - a_3b_3) \\ &+(a_0b_1 + a_1b_0 + a_2b_3 - a_3b_2)\,\boldsymbol{i} \\ &+(a_0b_2 - a_1b_3 + a_2b_0 + a_3b_1)\,\boldsymbol{j} \\ &+(a_0b_3 + a_1b_2 - a_2b_1 + a_3b_0)\,\boldsymbol{k} \end{aligned} \quad (2)$$

**Mapping from Quaternion-valued to Complex-Valued matrices:** We fix the complex subfield:

$$\mathbb{C}_{\boldsymbol{i}} = \{a + b\boldsymbol{i} : a, b \in \mathbb{R}\} \subset \mathbb{H}. \quad (3)$$

Then, every quaternion $q \in \mathbb{H}$ can be written uniquely as

$$q = \alpha + \beta\boldsymbol{j}, \qquad \alpha, \beta \in \mathbb{C}_{\boldsymbol{i}}, \quad (4)$$

using the relation $\mathbf{j}\alpha = \bar{\alpha}\mathbf{j}$ for $\alpha \in \mathbb{C}_{\boldsymbol{i}}$, where $\bar{\cdot}$ denotes complex conjugation in $\mathbb{C}_{\boldsymbol{i}}$. This yields the $\mathbb{C}_{\boldsymbol{i}}$-pair representation:

$$q \equiv (\alpha, \beta) \in \mathbb{C}_{\boldsymbol{i}} \times \mathbb{C}_{\boldsymbol{i}} \quad (5)$$

Let $q = (\alpha,\beta)$ and $p = (\gamma,\delta)$ with $\alpha,\beta,\gamma,\delta \in \mathbb{C}_{\mathbf{i}}$. The Hamilton product $qp$ corresponds to the pair product:

$$(\alpha,\beta) \odot (\gamma,\delta) = \big(\alpha\gamma - \beta\bar{\delta}, \alpha\delta + \beta\bar{\gamma}\big) \quad (6)$$

With this convention $1 \equiv (1,0)$, $\boldsymbol{i} \equiv (\boldsymbol{i}, 0)$, $\boldsymbol{j} \equiv (\mathbf{0}, 1)$ and $(\boldsymbol{i}, 0) \odot (\mathbf{0}, 1) = (0, \boldsymbol{i}) \equiv \mathbf{k}$.

Using the decomposition $q = \alpha + \beta\mathbf{j}$, we define the complex $2 \times 2$ matrix representation $M_{\mathbb{C}}: \mathbb{H} \to \mathbb{C}^{2\times 2}$ by

$$M_{\mathbb{C}}(q) = \begin{pmatrix} \alpha & \beta \\ -\overline{\beta} & \overline{\alpha} \end{pmatrix} \tag{7}$$

This map is an injective $\mathbb{R}$-algebra homomorphism: $M_{\mathbb{C}}(p+q) = M_{\mathbb{C}}(p) + M_{\mathbb{C}}(q)$ and $M_{\mathbb{C}}(pq) = M_{\mathbb{C}}(p)M_{\mathbb{C}}(q)$, and quaternion conjugation corresponds to conjugate transpose, $M_{\mathbb{C}}(\overline{q}) = M_{\mathbb{C}}(q)^*$.

For a quaternion matrix $A \in \mathbb{H}^{m\times n}$, write

$$A = A_1 + A_2\boldsymbol{j}, \qquad A_1, A_2 \in \mathbb{C}_{\boldsymbol{i}}^{m\times n} \tag{8}$$

Its complex adjoint representation is the block matrix

$$\chi(A) = \begin{pmatrix} A_1 & A_2 \\ -\overline{A_2} & \overline{A_1} \end{pmatrix} \in \mathbb{C}^{2m\times 2n} \tag{9}$$

Assembling the $2 \times 2$ blocks $M_{\mathbb{C}}(A_{ij})$ yields exactly $\chi(A)$ up to a fixed permutation of rows and columns. Hence $\chi$ is an $\mathbb{R}$-algebra isomorphism onto its image:

$$\chi(AB) = \chi(A)\chi(B) \tag{10}$$

$$\chi(A+B) = \chi(A) + \chi(B) \tag{11}$$

$$\chi(A^*) = \chi(A)^* \tag{12}$$

where $A^* = \bar{A}^{\top}$ is the quaternion conjugate transpose.

### *B. QUATERNION FOURIER TRANSFORMS*

The widely used discrete quaternion Fourier transforms depend on the selected basis vector and whether the transform is left-sided, right-sided, or two-sided [54]. We adopt the left-sided discrete quaternion Fourier transform (DQFT) [55], [56]. Let $\boldsymbol{\mu} \in \mathbb{H}$ be a pure unit quaternion with $\|\boldsymbol{\mu}\| = 1$. For a sequence $x[n] \in \mathbb{H}$, $n = 0, \dots, N-1$, the left-sided DQFT with axis $\boldsymbol{\mu}$ is

$$X[u] = \sum_{n=0}^{N-1} e^{-\boldsymbol{\mu} 2\pi un/N}\, x[n], \quad u = 0, \dots, N-1, \tag{14}$$

$$x[n] = \frac{1}{N}\sum_{u=0}^{N-1} e^{\boldsymbol{\mu} 2\pi un/N}\, X[u], \quad n = 0, \dots, N-1 \tag{15}$$

In matrix form:

$$X = Q_{\boldsymbol{\mu}}^N x, \qquad \left(Q_{\boldsymbol{\mu}}^N\right)_{u,n} = e^{-\boldsymbol{\mu} 2\pi un/N} \tag{16}$$

We fix the axis $\boldsymbol{\mu} = \mathbf{i}$, aligned with $\mathbb{C}_{\mathbf{i}}$. Writing $x[n] = \alpha[n] + \beta[n]\mathbf{j}$ with $\alpha, \beta \in \mathbb{C}_{\mathbf{i}}$, and using that $e^{-\mathbf{i}\theta} \in \mathbb{C}_{\mathbf{i}}$ commutes with $\alpha[n]$ and $\beta[n]$, we obtain:

$$X[u] = \hat{\alpha}[u] + \hat{\beta}[u]\boldsymbol{j} \tag{17}$$

where $\hat{\alpha} = \mathrm{FFT}(\alpha)$ and $\hat{\beta} = \mathrm{FFT}(\beta)$ are standard complex DFTs. Thus:

For axis $\boldsymbol{\mu} = \mathbf{i}$, the DQFT of $x = \alpha + \beta\mathbf{j}$ reduces to a pair of complex FFTs in $\mathbb{C}_{\mathbf{i}}$:

$$\hat{\alpha} = FFT(\alpha) \tag{18}$$

$$\hat{\beta} = FFT(\beta) \tag{19}$$

$$X = \hat{\alpha} + \hat{\beta}\boldsymbol{j} \tag{20}$$

with an analogous statement for the inverse transform.

At the matrix level, the Fourier matrix satisfies

$$\chi(Q_{\boldsymbol{i}}^N) = \frac{1}{\sqrt{N}}\begin{pmatrix} F_N & 0 \\ 0 & \overline{F_N} \end{pmatrix} \tag{21}$$

where $F_N$ is the complex DFT matrix with imaginary unit $\mathbf{i}$. Hence, after applying $\chi(\cdot)$, the quaternion DQFT reduces to two coupled complex FFTs.

### *C. QUATERNION CIRCULANT AND BLOCK-CIRCULANT MATRICES*

#### 1) QUATERNION CIRCULANT CONVOLUTION AND CIRCULANT MATRICES

For quaternion sequences $h[n], x[n] \in \mathbb{H}$ of length $N$, we define left circulant convolution (indices modulo $N$) by

$$(h * x)[n] = \sum_{k=0}^{N-1} h\,[k]x[(n-k) \bmod N] \tag{22}$$

Given a kernel $k_C = [c_0, \dots, c_{N-1}]^{\top} \in \mathbb{H}^N$, the associated left-convolution circulant matrix $C \in \mathbb{H}^{N\times N}$ has entries

$$C_{i,j} = c_{(i-j) \bmod N}, \qquad i,j = 0, \dots, N-1 \tag{23}$$

so that $Cx$ implements left circulant convolution with kernel $k_C$. Equivalently,

$$C = \sum_{n=0}^{N-1} c_n\, \tilde{P}^n \tag{24}$$

where $\tilde{P} \in \mathbb{R}^{N\times N}$ is the cyclic permutation matrix.

In the complex case, circulant matrices are fully diagonalized by the DFT: $F_N C F_N^{-1} = \mathrm{diag}(\hat{c})$. For quaternion circulant matrices, non-commutativity breaks this diagonalization. Pan and Ng [36] showed that, with respect to the quaternion Fourier matrix $Q_{\mathbf{i}}^N$, circulant matrices admit a block-diagonal form with $1 \times 1$ and $2 \times 2$ quaternion blocks, but there is no single quaternion Fourier matrix that simultaneously diagonalizes all circulant matrices as in the complex case [37].

#### 2) BLOCK-CIRCULANT QUATERNION MATRICES

Let $B$ be the number of blocks and $k$ the block size. A quaternion matrix $W \in \mathbb{H}^{(Bk)\times(Bk)}$ is block-circulant if it can be partitioned into $B \times B$ blocks $C_b \in \mathbb{H}^{k\times k}$ such that

$$W = \begin{pmatrix} C_0 & C_{B-1} & \cdots & C_1 \\ C_1 & C_0 & \cdots & C_2 \\ \vdots & \vdots & \ddots & \vdots \\ C_{B-1} & C_{B-2} & \cdots & C_0 \end{pmatrix} \tag{25}$$

$$C_b = circ\left(k_C^{(b)}\right) \tag{26}$$

where $C_b \in \mathbb{H}^{k\times k}$ are arbitrary quaternion-valued blocks.

Reshaping $x \in \mathbb{H}^{Bk}$ into blocks $x^{(b)} \in \mathbb{H}^k, b = 0, \dots, B-1$, the action of $W$ can be written as:

$$y^{(i)}[r] = \sum_{b=0}^{B-1}\sum_{s=0}^{k-1} C_b\,[r,s]x^{(i-b) \bmod B}[s] \tag{27}$$

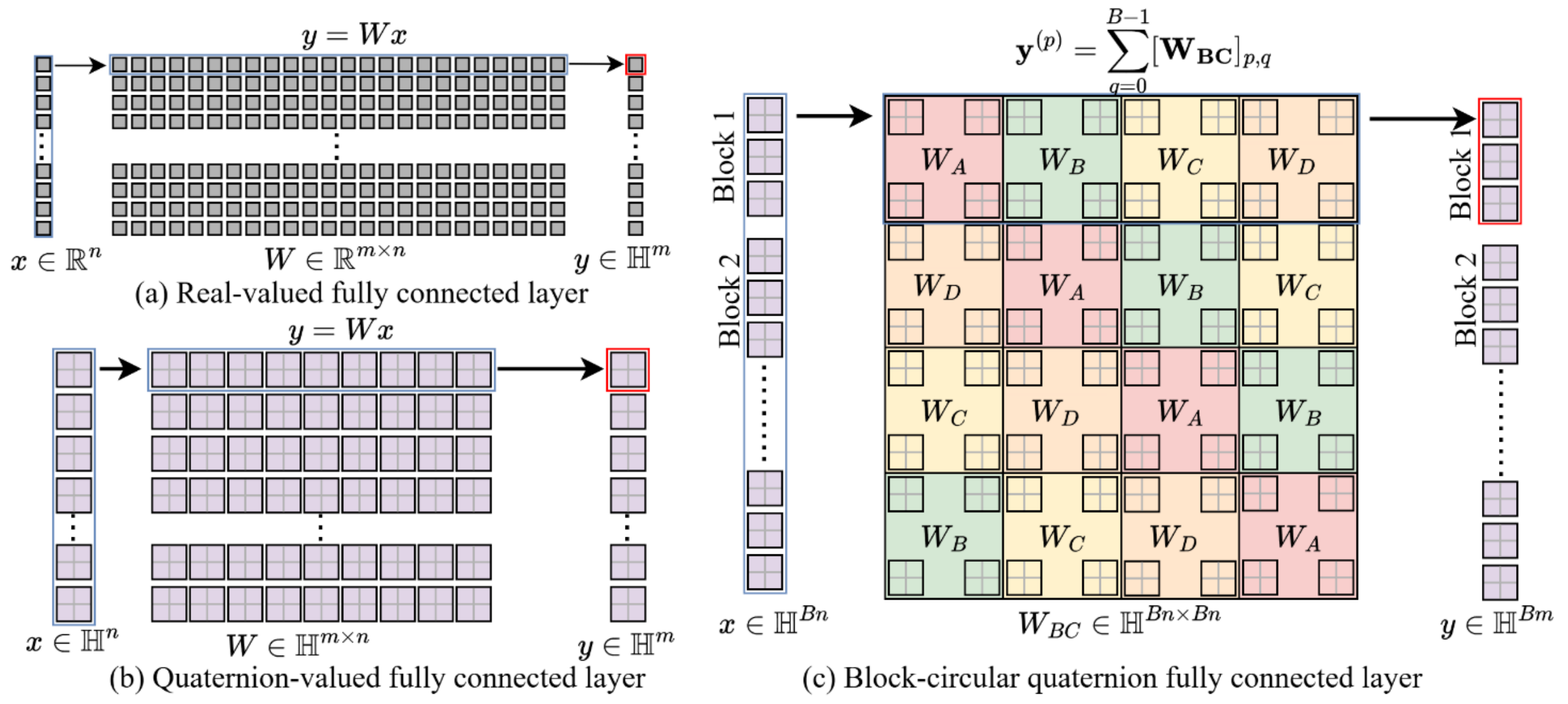

***FIGURE 1.*** **Comparison of linear fully connected layers without bias.(a) Real-valued fully connected layer: $x \in \mathbb{R}^n$ is mapped to $y \in \mathbb{R}^m$ by $y = Wx$.(b) Quaternion-valued fully connected layer: $x \in \mathbb{H}^n$, $W \in \mathbb{H}^{m\times n}$, $y \in \mathbb{H}^m$, and each quaternion output is $y_i = \sum_j W_{ij} x_j$, where multiplication is the (left) Hamilton product. (c) Block-circulant quaternion fully connected layer: the quaternion vector is partitioned into $B$ blocks $x^{(q)}$ and the weight matrix $W_{BC} \in \mathbb{H}^{(Bm)\times(Bn)}$ is block-circulant in the block index (constructed by cyclic shifts of a small set of quaternion submatrices). Each output block is computed as $y^{(p)} = \sum_{q=0}^{B-1}[W_{BC}]_{p,q} x^{(q)}$, where multiplication is the Hamilton product.**

which can be interpreted as a two-dimensional convolution over the block index and the within-block index when each $C_b$ itself has a circulant structure. We define the block Fourier operator as

$$Q_B = Q_{\boldsymbol{i}}^B \otimes I_k \in \mathbb{H}^{(Bk)\times(Bk)} \tag{28}$$

where $\otimes$ denotes the Kronecker product and $I_k$ is $k \times k$ identity matrix. Using the polynomial representation $W = \sum_{b=0}^{B-1} \tilde{P}_B^b \otimes C_b$ with $\tilde{P}_B \in \mathbb{R}^{B\times B}$ the cyclic permutation matrix, and the fact that $Q_{\mathbf{i}}^B$ diagonalizes $\tilde{P}_B$, we obtain:

$$Q_B W Q_B^{-1} = \bigoplus_{u=0}^{B-1} \tilde{C}_u \tag{29}$$

$$\tilde{C}_u = \sum_{b=0}^{B-1} \omega_B^{ub}\, C_b \tag{30}$$

$$\omega_B = e^{-\boldsymbol{i}2\pi/B} \tag{31}$$

where $\oplus$ denotes the block direct sum.

The matrix $W$ is thus block-diagonalized into $B$ independent $k \times k$ quaternion blocks $\tilde{C}_u$. Unlike the complex case, these blocks are not diagonal in general due to non-commutativity, but this block structure is sufficient for efficient FFT-based algorithms when combined with the complex adjoint $\chi(\cdot)$.

### *D. BLOCK-CIRCULANT QUATERNION LAYERS*

This section specifies the structured quaternion layers used in EdgeLDR by embedding the quaternion circulant operators into trainable neural-network mappings.

Our design imposes circulant parameter sharing within small channel groups. Concretely, the weight operator is organized as a block matrix whose blocks are quaternion circulant matrices generated by short quaternion kernels. The block size k controls the compression–expressivity trade-off: larger k yields stronger parameter tying (higher compression) and longer FFTs; $(k = 1)$ reduces to an unstructured quaternion layer.

#### 1) FULLY CONNECTED LAYER

Let $x \in \mathbb{H}^n$ and $y \in \mathbb{H}^m$. Choose a block count $B$ such that $B \mid n$ and $B \mid m$, and set $n = Bd_{\text{in}}$ and $m = Bd_{\text{out}}$. Split the activations into channel blocks $x = [x^0, \dots, x^{B-1}]$ with $x^q \in \mathbb{H}^{d_{\text{in}}}$ and $y = [y^0, \dots, y^{B-1}]$ with $y^p \in \mathbb{H}^{d_{\text{out}}}$. Define the mod index $[t]_B = t \bmod B$ and the index set $[B] = \{0, \dots, B-1\}$.

EdgeLDR parameterizes the weight $W \in \mathbb{H}^{m\times n}$ as a $B \times B$ block matrix with dense quaternion blocks. It stores only the first block row as $B$ learnable quaternion blocks $K_s \in \mathbb{H}^{d_{\text{out}}\times d_{\text{in}}}$ for $s \in [B]$ and sets

$$W_{p,q} = K_{[q-p]_B}, \qquad p, q \in [B] \tag{32}$$

The forward mapping is

$$y^p = \sum_{q\in[B]} \mathrm{K}_{[q-p]_B} x^q + b^p, p \in [B] \tag{33}$$

where $b^p \in \mathbb{H}^{d_{\text{out}}}$ is an optional bias block and products use left Hamilton multiplication inside quaternion matrix–vector products.

Each $K_s$ is stored by its four real components $K_s^{\mathrm{r}}, K_s^{\mathrm{i}}, K_s^{\mathrm{j}}, K_s^{\mathrm{k}} \in \mathbb{R}^{d_{\text{out}}\times d_{\text{in}}}$, realized as learnable tensors of shape $B \times d_{\text{out}} \times d_{\text{in}}$ for r, i, j, k. At runtime, these are expanded into dense real matrices by block-circulant shifts and applied using the standard Hamilton packing so the layer reduces to one real matrix multiply plus optional bias.

.

2) CONVOLUTION LAYER

Let $X \in \mathbb{H}^{n\times H\times W}$ and $Y \in \mathbb{H}^{m\times H'\times W'}$ denote the input and output feature maps in quaternion channels. We choose a block count $B$ such that $n = Bd_{\text{in}}$ and $m = Bd_{\text{out}}$. We split channels into blocks $X = [X^0, \dots, X^{B-1}]$ with $X^q \in \mathbb{H}^{d_{\text{in}}\times H\times W}$ and $Y = [Y^0, \dots, Y^{B-1}]$ with $Y^p \in \mathbb{H}^{d_{\text{out}}\times H'\times W'}$. Use $[t]_B = t \bmod B$ and $[B] = \{0, \dots, B-1\}$.

EdgeLDR uses standard spatial convolution while constraining channel mixing by a block-circulant pattern over the channel blocks. For kernel size $K_h \times K_w$, it stores only the first block row as $B$ learnable quaternion kernel blocks $K_s \in \mathbb{H}^{d_{\text{out}}\times d_{\text{in}}\times K_h\times K_w}$ for $s \in [B]$ and defines the full kernel bank by

$$W_{p,q} = K_{[q-p]_B}, \qquad p, q \in [B] \quad (34)$$

with $W_{p,q} \in \mathbb{H}^{d_{\text{out}}\times d_{\text{in}}\times K_h\times K_w}$.

The block output is

$$y^p = \sum_{q\in[B]} \mathrm{K}_{[q-p]_B} \star X^q + b^p, \qquad p \in [B] \quad (35)$$

where $\star$ is the 2D quaternion convolution over spatial offsets using left Hamilton products for channel mixing, and $b^p \in \mathbb{H}^{d_{\text{out}}}$ is an optional bias block. Stride, padding, and dilation follow the standard Conv2D definition.

Each $K_s$ is stored by real components $K_s^{\text{r}}, K_s^{\text{i}}, K_s^{\text{j}}, K_s^{\text{k}} \in \mathbb{R}^{d_{\text{out}}\times d_{\text{in}}\times K_h\times K_w}$, realized as learnable tensors of shape $B \times d_{\text{out}} \times d_{\text{in}} \times K_h \times K_w$ for r, i, j, k. These are expanded by block-circulant shifts into dense r, i, j, k kernels and applied with a single real Conv2D via Hamilton packing.

Consider a dense quaternion convolution with $n$input and $m$output quaternion channels and kernel size $K_h \times K_w$. The unconstrained layer contains $m\,n\,K_hK_w$quaternion weights, i.e., $4\,m\,n\,K_hK_w$ real degrees of freedom (one real component for each of $r, i, j, k$). Under the proposed channel partition $n = Bd_{\text{in}}$and $m = Bd_{\text{out}}$, the BC-QConv layer stores only the first block row $\{K_s\}_{s=0}^{B-1}$ with $K_s \in \mathbb{H}^{d_{\text{out}}\times d_{\text{in}}\times K_h\times K_w}$. The resulting trainable weight budget is therefore

$$B\, d_{\text{out}}\, d_{\text{in}}\, K_hK_w = (m\,n/B)\, K_hK_w$$

quaternion parameters, i.e., $4(m\,n/B)K_hK_w$ real parameters, plus an optional bias of $m$ quaternions. Thus, for fixed $(m, n, K_h, K_w)$, the block factor $B$ reduces weight storage by approximately a factor of $B$ relative to a dense quaternion convolution. When comparing against an "equivalent" real convolution operating on $4n$ real input channels and producing $4m$ real output channels, the weight reduction is approximately $4B$ .

An important note is that this compression is achieved without changing the spatial receptive field or the 2d convolution operator definition: stride/padding/dilation and spatial kernel size remain standard, while the channel-mixing degrees of freedom are tied cyclically across blocks.

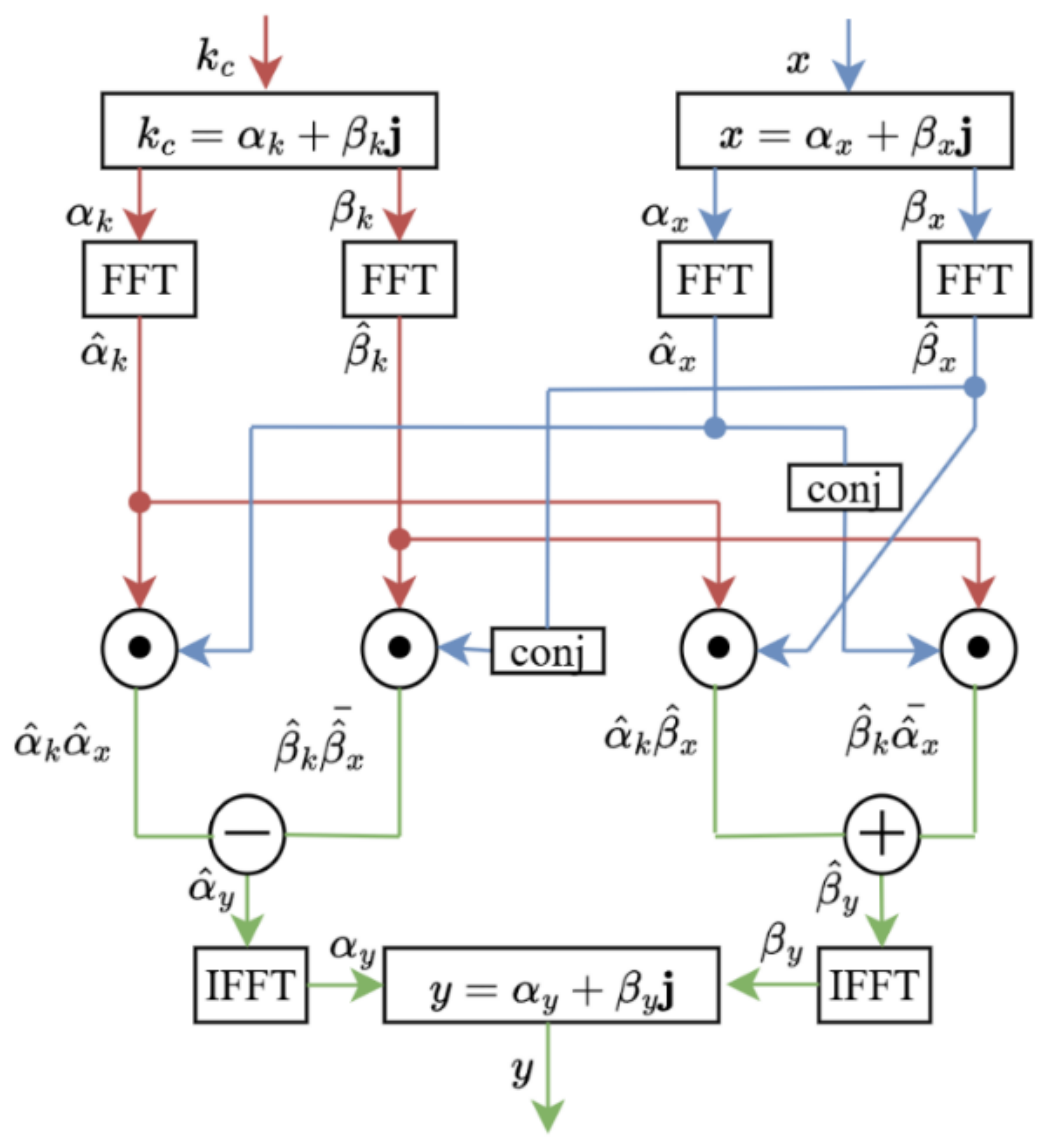

**FIGURE 2. Computational flow diagram for FFT-based quaternion circulant matrix-vector multiplication. The diagram illustrates the efficient algorithm for quaternion circulant convolution used in EdgeLDR layers. Both the quaternion kernel and input vector are first decomposed into pairs of complex components. Each component undergoes a standard complex FFT to transform into the frequency domain. In the frequency domain, the quaternion multiplication is realized through a specific pattern of complex multiplications, conjugations, additions, and subtractions that respects the non-commutative Hamilton product rules. The left branch combines terms with subtraction while the right branch uses addition. Finally, inverse FFTs convert the frequency-domain results back to the spatial domain. Red arrows indicate the kernel path, blue arrows the input path, and green arrows the output path.**

### *E. ACCELERATED COMPUTATION*

The key algorithmic ingredient in EdgeLDR layers is the reduction of quaternion circulant matrix-vector products to pairs of complex FFTs and simple frequency-wise algebra via the bicomplex representation (Figure 2). Consider $y = Cx$, where $C \in \mathbb{H}^{k\times k}$ is the circulant matrix associated with kernel $k_C \in \mathbb{H}^k$, then using (3) we can write:

$$x = \alpha_x + \beta_x \boldsymbol{j} \quad (36)$$

$$k_C = \alpha_k + \beta_k \boldsymbol{j} \quad (37)$$

with $\alpha_x, \beta_x, \alpha_k, \beta_k \in \mathbb{C}_{\mathbf{i}}^k$. Let's compute complex FFTs:

$$\hat\alpha_x = FFT(\alpha_x) \quad (38)$$

$$\hat\beta_x = FFT(\beta_x) \quad (39)$$

$$\hat\alpha_k = FFT(\alpha_k) \quad (40)$$

$$\hat\beta_k = FFT(\beta_k) \quad (41)$$

In the DQFT domain, the convolution theorem gives:

$$Y[u] = K[u]X[u] \quad (42)$$

$$X[u] = \hat\alpha_x[u] + \hat\beta_x[u]\boldsymbol{j} \quad (43)$$

$$K[u] = \hat\alpha_k[u] + \hat\beta_k[u]\boldsymbol{j} \quad (44)$$

ALGORITHM 1. FFT-BASED QUATERNION CIRCULANT PRODUCT

**Require:** Quaternion kernel $k_C \in \mathbb{H}^k$; vector $x \in \mathbb{H}^k$.
**Ensure:** $y = Cx$ with $C = \mathrm{circ}(k_C)$.

Choose subfield $\mathbb{C}_\mathbf{i}$. Write $x = \alpha_x + \beta_x \mathbf{j}$.
Write $k_C = \alpha_k + \beta_k \mathbf{j}$.
$\hat{\alpha}_x \leftarrow \mathrm{FFT}(\alpha_x)$; $\hat{\beta}_x \leftarrow \mathrm{FFT}(\beta_x)$.
$\hat{\alpha}_k \leftarrow \mathrm{FFT}(\alpha_k)$; $\hat{\beta}_k \leftarrow \mathrm{FFT}(\beta_k)$.
**for** $u = 0, \dots, k-1$ **do**
  $\hat{\alpha}_y[u] = \hat{\alpha}_k[u]\hat{\alpha}_x[u] - \hat{\beta}_k[u]\overline{\hat{\beta}_x[u]}$.
  $\hat{\beta}_y[u] = \hat{\alpha}_k[u]\hat{\beta}_x[u] + \hat{\beta}_k[u]\overline{\hat{\alpha}_x[u]}$.
**end for**
$\alpha_y \leftarrow \mathrm{IFFT}(\hat{\alpha}_y)$; $\beta_y \leftarrow \mathrm{IFFT}(\hat{\beta}_y)$.
$y \leftarrow \alpha_y + \beta_y \mathbf{j}$.

$$Y[u] = \hat{\alpha}_y[u] + \hat{\beta}_y[u]\boldsymbol{j} \tag{45}$$

with

$$\hat{\alpha}_y[u] = \hat{\alpha}_k[u]\hat{\alpha}_x[u] - \hat{\beta}_k[u]\overline{\hat{\beta}_x[u]} \tag{46}$$

$$\hat{\beta}_y[u] = \hat{\alpha}_k[u]\hat{\beta}_x[u] + \hat{\beta}_k[u]\overline{\hat{\alpha}_x[u]} \tag{47}$$

where overline denotes complex conjugation in $\mathbb{C}_i$. Finally, inverse FFTs yield:

$$\alpha_y = IFFT(\hat{\alpha}_y) \tag{48}$$

$$\beta_y = IFFT(\hat{\beta}_y) \tag{49}$$

$$y = \alpha_y + \beta_y \boldsymbol{j} \tag{50}$$

This computation uses four complex FFTs of length $k$, plus $O(k)$ complex multiplications per frequency. Compared to naive $O(k^2)$ quaternion convolution, the cost is $O(k\log k)$ in the circulant dimension, matching real/complex FFT-based convolutions up to a constant factor from quaternion arithmetic.

## IV. EXPERIMENTS

This section evaluates EdgeLDR on two representative workloads that reflect common edge-vision constraints: compact convolutional networks for 32×32 RGB classification and compact Transformer models for hyperspectral image (HSI) classification. All experiments are conducted on a workstation equipped with an AMD Ryzen 9 9950X3D CPU, an NVIDIA RTX 5080 GPU, and 64 GB of DDR5 memory using PyTorch 2.9 [57]. For RGB benchmarks we report top-1 test accuracy and parameter counts. For hyperspectral benchmarks we report mean and standard deviation over five random seeds for Overall Accuracy (OA), Average Accuracy (AA), and the Kappa coefficient κ, together with model size and measured inference latency.

### A. RGB IMAGE CLASSIFICATION

We evaluate on CIFAR-10, CIFAR-100, and SVHN [58], [59]. CIFAR-10/100 follow the standard 50k/10k train/test split with per-channel normalization, random horizontal flips, and random crops with padding. SVHN uses the official training and test splits with per-channel normalization and random translations; horizontal flips are omitted. All results in Table I use a ResNet-50 backbone adapted to 32×32 inputs. The network begins with a 3×3 convolution with stride 1 and no max pooling, followed by four residual stages, where stages 2–4 start with stride-2 downsampling and each residual block contains two 3×3 convolutions. The Real baseline uses standard real-valued convolutions and a real fully connected classifier. The Quaternion baseline replaces these layers with quaternion convolutions and quaternion linear layers and encodes each RGB pixel as a pure quaternion with components aligned to the color channels. PHM models replace the same set of layers with parameterized hypercomplex multiplication, using hypercomplex dimension n equal to 2 or 4 [60]. Real block-circulant models impose CirCNN-style block-circulant structure on each weight tensor, using block size g equal to 2 or 4 [17]. EdgeLDR combines both design principles by imposing the block-circulant constraint directly in the quaternion domain; the configuration reported in Table I corresponds to block size g=2.

All networks are trained from scratch using stochastic gradient descent with momentum 0.9, weight decay 5×10⁻⁴, batch size 128, and cross-entropy loss. The learning rate starts at 0.1 and follows cosine annealing over 200 epochs.

Table III summarizes the accuracy–parameter trade-offs. The dense real ResNet-50 reaches 94.96% on CIFAR-10, 78.06% on CIFAR-100, and 96.29% on SVHN with 23.5M parameters. When the constraint is mild, structured compression can preserve accuracy well: BC-2 reduces the parameter count to 11.8M while remaining close to the real baseline, and it slightly improves CIFAR-100 by 0.66 points, suggesting that moderate parameter tying can provide an implicit regularization effect in this setting. PHM-2 matches the same 11.8M budget and achieves the best SVHN accuracy in Table I at 96.45%. At stronger compression, the differences between structured parameterizations become more pronounced.

TABLE I
RGB IMAGE CLASSIFICATION WITH RESNET-50 BACKBONE

| Model | #Parameters (M)↓ | CIFAR-10↑ | CIFAR-100↑ | SVHN↑ |
|---|---|---|---|---|
| ResNet50 Real | 23.5 | 94.96 | 78.06 | 96.29 |
| ResNet50 BC-2 | 11.8 | 94.52 | 78.72 | 96.06 |
| ResNet50 PHM-2 | 11.8 | 93.89 | 78.63 | 96.45 |
| ResNet50 Quaternion | 5.9 | 93.65 | 78.16 | 96.30 |
| ResNet50 BC-4 | 5.9 | 93.82 | 75.78 | 95.78- |
| ResNet50 PHM-4 | 6.0 | 93.63 | 77.52 | 95.33 |
| ResNet50 EdgeLDR | 3.0 | 94.21 | 76.96 | 96.33 |

Dense quaternion-valued neural network reduces parameters from 23.5M to 5.9M, which is approximately a 4× reduction, while maintaining competitive accuracy across datasets. In contrast, enforcing a real block-circulant constraint at a similar parameter budget, as in BC-4 with 5.9M parameters, yields a noticeable drop on CIFAR-100 to 75.78%, indicating that aggressive structured tying in the real domain can underfit the more fine-grained 100-class task.

EdgeLDR delivers the most aggressive compression in Table III. With 3.0M parameters, EdgeLDR uses 7.83× fewer parameters than the dense real baseline and 1.97× fewer than the dense quaternion baseline, while achieving 94.21% on CIFAR-10, 76.96% on CIFAR-100, and 96.33% on SVHN. Relative to the real model, the CIFAR-10 and CIFAR-100 gaps are 0.75 and 1.10 points, respectively, and the SVHN accuracy is slightly higher. Importantly, EdgeLDR improves on BC-4 despite using roughly half as many parameters, gaining 0.39 points on CIFAR-10 and 1.18 points on CIFAR-100. This consistent advantage over purely real structured compression at comparable or smaller budgets supports the central premise of EdgeLDR: coupling structured block-circulant sharing with quaternion channel mixing helps retain representational capacity under tight parameter constraints.

### B. HYPERSPECTRAL IMAGE CLASSIFICATION

We evaluate on Houston 2013 and Pavia University. For each labeled pixel, a 7×7 spatial neighborhood is extracted using mirror padding. A compact spectral–spatial input is formed by stacking three adjacent spectral bands, consisting of the center band and its two nearest neighbors. The Real baseline is a SpectralFormer-style spectral Transformer using the cross-attention fusion variant [61]. The model uses embedding dimension 64, depth 5, four attention heads, MLP hidden dimension 8, dropout 0.1, followed by a linear classifier. To isolate the impact of the linear operator parameterization, the Transformer topology is held fixed and all linear projections in the embedding, self-attention, and MLP blocks are replaced with their corresponding variants. We compare dense quaternion linear layers, real block-circulant layers with block size g equal to 2, 4, and 8, PHM layers with hypercomplex dimension n equal to 2, 4, and 8, and EdgeLDR layers with block size g equal to 2 and 4. We also evaluate post-hoc compression of the Real baseline using dynamic post-training quantization of linear layers and unstructured magnitude pruning at 25% and 50%, followed by 10 epochs of fine-tuning. For pruned models, the parameter counts in Tables II and III reflect the remaining nonzero weights.

All models are trained from scratch for 1000 epochs using Adam with learning rate $5\times10^{-4}$, weight decay $5\times10^{-3}$, and batch size 64. A step schedule decays the learning rate by a factor of 0.9 every tenth of the training epochs. We report mean and standard deviation over five random seeds for OA, AA, and κ. Inference latency is measured using batch size 64 with 10 warm-up batches followed by 100 timed batches on both GPU and CPU; quantized models are evaluated on CPU only.

Results on Houston 2013: Table II shows that the Real baseline reaches 99.15 ± 0.32 OA with 241,496 parameters. EdgeLDR improves both accuracy and compactness. EdgeLDR-2 achieves the best performance at 99.86 ± 0.13 OA and κ=0.9985 ± 0.0014 with 151,896 parameters. This corresponds to a 37% reduction in trainable parameters and a 0.34 MB reduction in model size, together with a 0.71 point OA gain over the Real baseline. At an equal parameter budget, the real block-circulant baseline BC-8 attains 99.65 ± 0.00 OA, so EdgeLDR-2 provides a 0.21 point OA gain without increasing model size. EdgeLDR-4 further compresses the model to 145,496 parameters and maintains 99.79 ± 0.13 OA. Dynamic quantization preserves the Real model's accuracy while reducing its size, but it does not surpass the best structured variants. Unstructured pruning reduces the number of nonzero weights but consistently decreases accuracy and does not materially reduce the stored model size in the dense format reflected by the size column.

TABLE IV
HYPERSPECTRAL IMAGE CLASSIFICATION ON HOUSTON (MEAN AND STD OVER 5 SEEDS)
BEST ACCURACY IN EACH DATASET COLUMN IS SHOWN IN BOLD; THE SECOND-BEST IS UNDERLINED.

| Model | #Params↓ | Size (MB)↓ | OA (%)↑ | AA (%)↑ | κ↑ | GPU Latency (ms/batch)↓ | CPU Latency (ms/batch)↓ |
|---|---|---|---|---|---|---|---|
| Real | 241496 | 0.95 | 99.15 ± 0.32 | 99.17 ± 0.32 | 0.9909 ± 0.0035 | 2.60 ± 0.03 | 108.08 ± 2.39 |
| Real quantized | 241496 | 0.67 | 99.12 ± 0.34 | 99.13 ± 0.33 | 0.9905 ± 0.0036 | N/A | 103.06 ± 1.61 |
| Real pruned 25% | 215656 | 0.95 | 98.80 ± 0.48 | 98.81 ± 0.47 | 0.9871 ± 0.0051 | 2.50 ± 0.06 | 110.20 ± 1.35 |
| Real pruned 50% | 189816 | 0.95 | 98.90 ± 0.36 | 98.92 ± 0.36 | 0.9882 ± 0.0039 | 2.47 ± 0.02 | 109.90 ± 1.22 |
| Quaternion | 164696 | 0.68 | 99.65 ± 0.19 | 99.65 ± 0.19 | 0.9962 ± 0.0021 | 5.29 ± 0.02 | 108.99 ± 1.02 |
| BC-2 | 190296 | 0.77 | 99.47 ± 0.11 | 99.48 ± 0.12 | 0.9943 ± 0.0012 | 5.88 ± 0.02 | 238.94 ± 3.36 |
| BC-4 | 164696 | 0.66 | 99.72 ± 0.09 | 99.72 ± 0.08 | 0.9970 ± 0.0009 | 5.86 ± 0.06 | 187.37 ± 1.73 |
| BC-8 | <u>151896</u> | <u>0.61</u> | 99.65 ± 0.00 | 99.66 ± 0.00 | 0.9962 ± 0.0000 | 5.89 ± 0.04 | 171.47 ± 2.70 |
| PHM-2 | 191424 | 0.76 | 99.65 ± 0.30 | 99.66 ± 0.29 | 0.9962 ± 0.0032 | 4.02 ± 0.04 | 109.55 ± 0.53 |
| PHM-4 | 167000 | 0.67 | 99.65 ± 0.19 | 99.65 ± 0.19 | 0.9962 ± 0.0021 | 4.06 ± 0.04 | 110.50 ± 1.46 |
| PHM-8 | 163608 | 0.66 | 99.75 ± 0.21 | 99.76 ± 0.21 | 0.9973 ± 0.0023 | 4.07 ± 0.08 | 109.53 ± 0.35 |
| EdgeLDR-2 (our) | <u>151896</u> | <u>0.61</u> | **99.86 ± 0.13** | **99.86 ± 0.13** | **0.9985 ± 0.0014** | 11.81 ± 0.07 | 192.32 ± 1.19 |
| EdgeLDR-4 (our) | **145496** | **0.58** | <u>99.79 ± 0.13</u> | <u>99.79 ± 0.13</u> | <u>0.9977 ± 0.0014</u> | 11.13 ± 0.04 | 165.76 ± 1.00 |

TABLE V
HYPERSPECTRAL IMAGE CLASSIFICATION ON INDIAN PAVIA UNIVERSITY (MEAN AND STD OVER 5 SEEDS)
BEST ACCURACY IN EACH DATASET COLUMN IS SHOWN IN BOLD; THE SECOND-BEST IS UNDERLINED.

| Model | #Params↓ | Size (MB)↓ | OA (%)↑ | AA (%)↑ | $\kappa$↑ | GPU Latency (ms/batch)↓ | CPU Latency (ms/batch)↓ |
|---|---|---|---|---|---|---|---|
| Real | 177105 | 0.72 | 99.37 ± 0.18 | 99.11 ± 0.23 | 0.9914 ± 0.0025 | 2.41 ± 0.04 | 75.65 ± 0.43 |
| Real quantized | 177105 | 0.42 | 99.35 ± 0.18 | 99.10 ± 0.23 | 0.9911 ± 0.0025 | N/A | 71.69 ± 1.21 |
| Real pruned 25% | 151361 | 0.70 | 99.19 ± 0.20 | 99.03 ± 0.20 | 0.9888 ± 0.0027 | 2.45 ± 0.05 | 75.92 ± 0.57 |
| Real pruned 50% | 125617 | 0.70 | 99.11 ± 0.44 | 99.05 ± 0.33 | 0.9878 ± 0.0061 | 2.43 ± 0.06 | 75.69 ± 0.90 |
| Quaternion | 100305 | 0.43 | 99.03 ± 0.11 | 98.66 ± 0.25 | 0.9867 ± 0.0015 | 5.09 ± 0.06 | 78.08 ± 1.01 |
| BC-2 | 125905 | 0.51 | 99.30 ± 0.11 | 99.07 ± 0.14 | 0.9904 ± 0.0015 | 5.61 ± 0.04 | 177.99 ± 1.35 |
| BC-4 | 100305 | 0.41 | 99.37 ± 0.07 | 99.26 ± 0.06 | 0.9914 ± 0.0010 | 5.67 ± 0.05 | 135.55 ± 0.96 |
| BC-8 | 87505 | 0.37 | 99.41 ± 0.13 | 99.21 ± 0.19 | 0.9919 ± 0.0018 | 5.58 ± 0.06 | 124.33 ± 1.77 |
| PHM-2 | 127033 | 0.53 | 99.15 ± 0.21 | 98.95 ± 0.23 | 0.9884 ± 0.0028 | 3.84 ± 0.03 | 79.07 ± 0.58 |
| PHM-4 | 102609 | 0.43 | 99.32 ± 0.21 | 99.17 ± 0.29 | 0.9908 ± 0.0028 | 3.81 ± 0.02 | 79.08 ± 0.36 |
| PHM-8 | 99217 | 0.42 | 99.53 ± 0.16 | 99.32 ± 0.19 | 0.9936 ± 0.0022 | 3.81 ± 0.02 | 80.54 ± 0.62 |
| EdgeLDR-2 (our) | 87505 | 0.37 | 99.41 ± 0.05 | 99.24 ± 0.08 | 0.9919 ± 0.0007 | 11.89 ± 0.38 | 139.49 ± 0.85 |
| EdgeLDR-4 (our) | **81105** | **0.34** | **99.62 ± 0.09** | **99.48 ± 0.10** | **0.9948 ± 0.0012** | 11.11 ± 0.09 | 125.11 ± 3.38 |

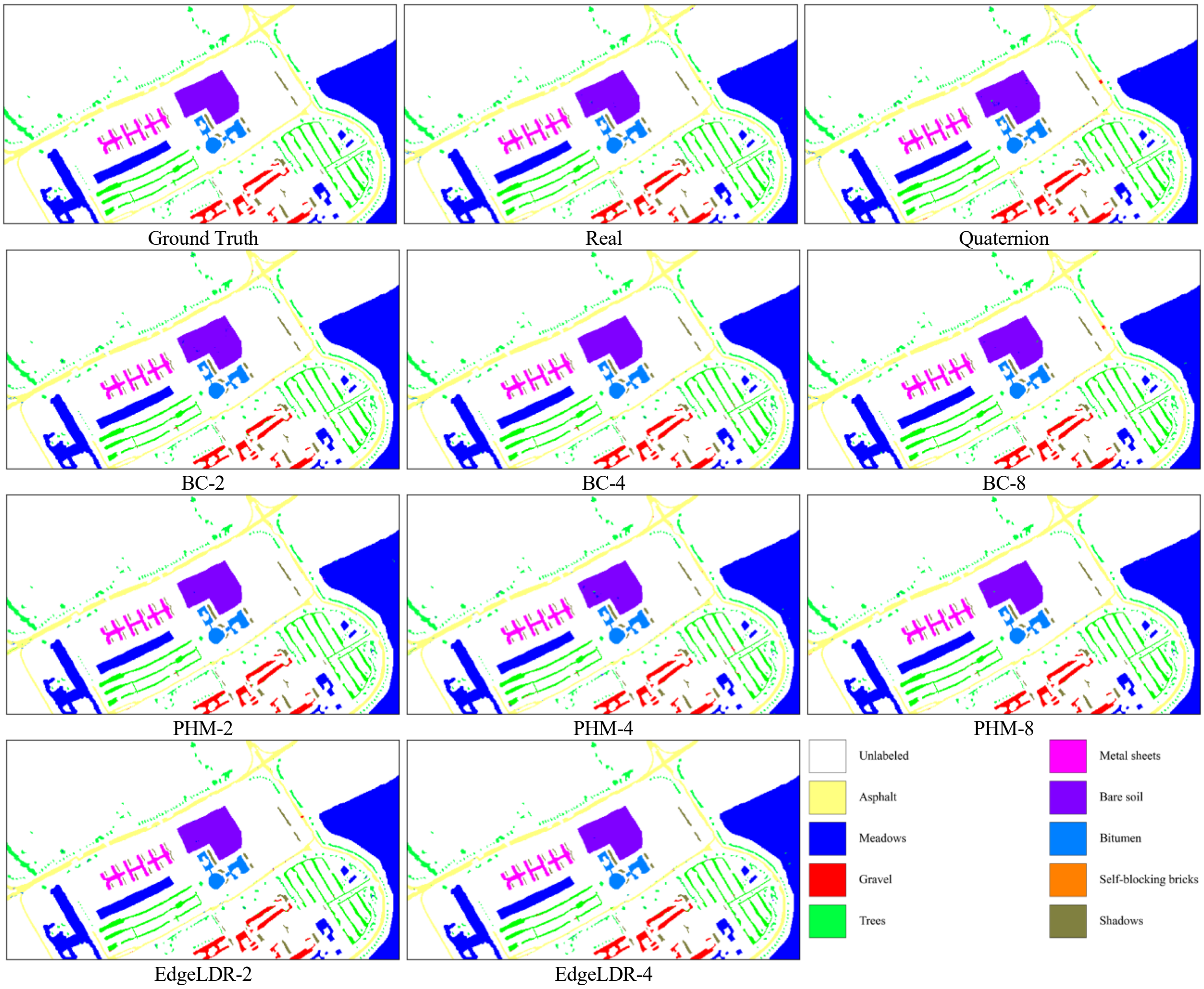

**FIGURE 3. Full-scene hyperspectral classification on Pavia University. Ground truth and predicted label maps are shown for the Real baseline and all structured/hypercomplex variants evaluated in Table V (Real, Quaternion, BC-2/4/8, PHM-2/4/8, and EdgeLDR-2/4). Predictions are masked to labeled pixels due to sparse annotation; unlabeled regions are left blank. The qualitative comparison illustrates that EdgeLDR variants preserve the global spatial layout and class boundaries while operating at a substantially reduced parameter budget.**

Results on Pavia University: Table III reports similarly high accuracies across all methods, reflecting a near-saturated regime. The Real baseline reaches 99.37 ± 0.18 OA with 177,105 parameters. EdgeLDR-4 attains the best OA at 99.62 ± 0.09 and the best κ at 0.9948 ± 0.0012 while using only 81,105 parameters. This is a 2.18× reduction in parameter count and a 0.38 MB reduction in model size relative to the Real baseline, together with a 0.25 point OA gain. EdgeLDR-2 matches the Real model's OA with roughly half the parameters. Fig. 3 visualizes full-scene predictions on Pavia University, masked to labeled pixels due to sparse annotation, and confirms that EdgeLDR produces predictions consistent with the strongest baselines while preserving the global spatial layout.

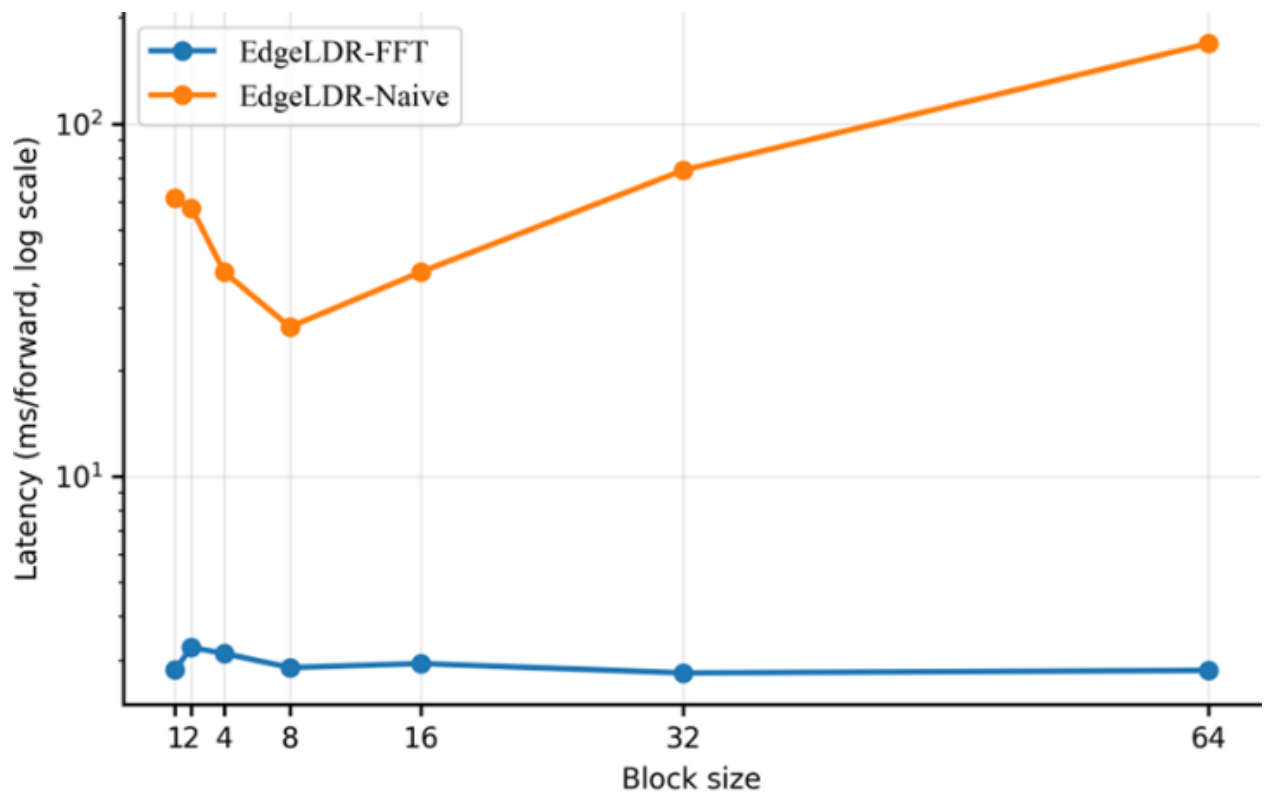

**FIGURE 4. FFT vs. naive runtime. Measured forward-pass latency for a 6-layer quaternion MLP using the same EdgeLDR parameterization but two implementations: EdgeLDR-FFT (FFT-based quaternion circulant products) and EdgeLDR-Naive (direct spatial-domain computation). The plot shows that FFT evaluation maintains near-constant latency across increasing block sizes, whereas the naive implementation becomes substantially slower as block size grows.**

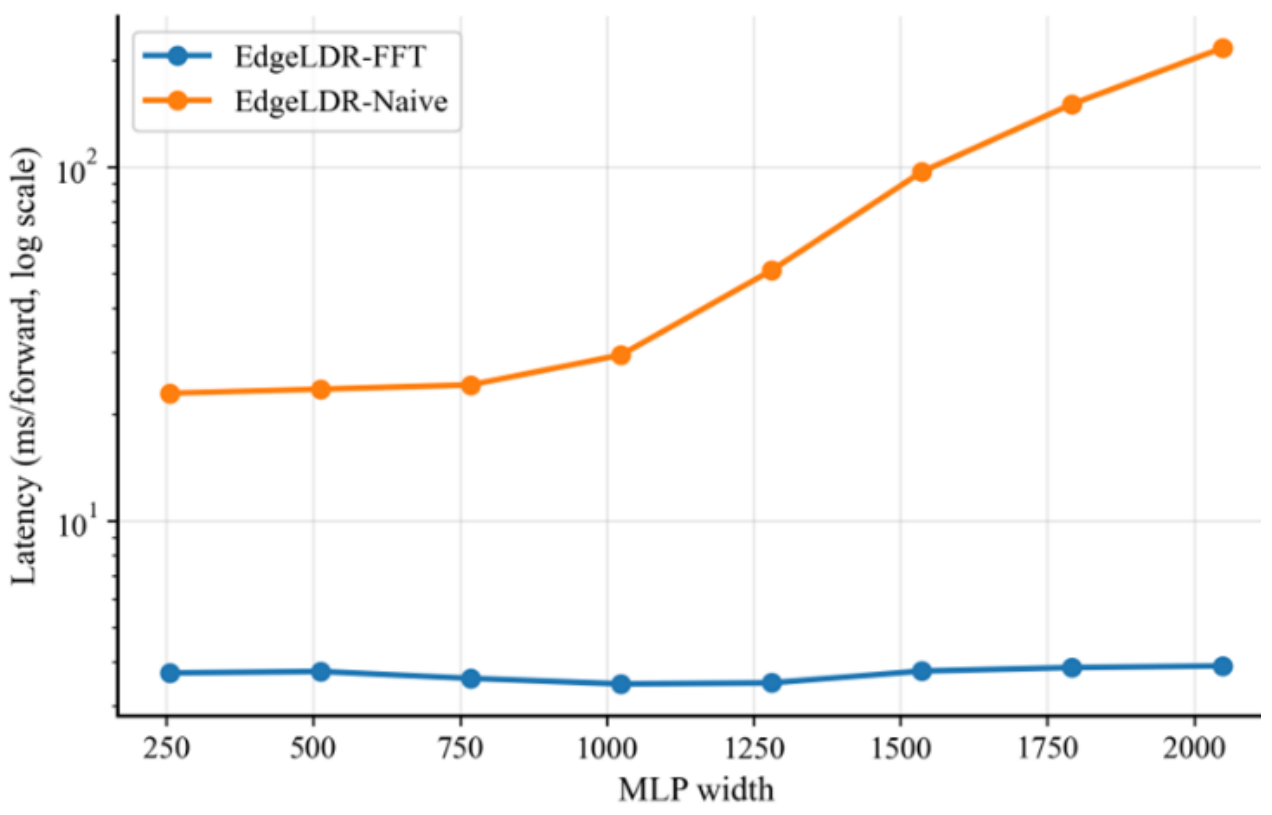

**FIGURE 5. FFT vs. naive runtime under a width sweep at fixed block size 8. Forward-pass latency for the same 6-layer quaternion MLP while varying model width from 256 to 2048. The FFT implementation scales smoothly with width and remains consistently faster than the naive spatial-domain realization, confirming that the FFT path is the practical mechanism for deploying EdgeLDR.**

Tables II and III show that parameter efficiency does not automatically translate into lower end-to-end latency under a straightforward PyTorch implementation. The Real baseline is the fastest on GPU, and PHM variants remain closest to Real in both GPU and CPU latency while still reducing parameters. Block-circulant and EdgeLDR variants incur higher latency, particularly on GPU where EdgeLDR is about 4–5× slower than the Real baseline for this Transformer configuration. On CPU, EdgeLDR is substantially slower than the Real and PHM baselines but is comparable to, and sometimes faster than, real block-circulant baselines at similar model size.Dynamic quantization provides the most consistent CPU-side size and latency improvement with minimal accuracy change, whereas unstructured pruning yields limited runtime benefit because sparsity is not exploited by dense kernels and the serialized model size changes only marginally.

### *C. ABLATIONS*

EdgeLDR relies on FFT-based evaluation of block-circulant quaternion operators. To isolate the computational impact of this design choice, we benchmark a random, untrained six-layer MLP with quaternion width 1024 and batch size 256 and compare the FFT-based implementation against a direct spatial-domain implementation under identical parameterization. Timing uses 50 warm-up iterations followed by 200 timed iterations on the GPU.

Fig. 4 varies the block size k over 1, 2, 4, 8, 16, 32, and 64. The FFT implementation remains in a narrow latency band between 2.77 ms and 3.27 ms per forward pass across the sweep, while the naive implementation ranges from 26.5 ms to 168.9 ms. In other words, FFT evaluation provides an empirical speedup of roughly 8× to 60× in this setting and is essential for making larger blocks computationally viable. Fig. 5 fixes k=8 and sweeps the MLP width from 256 to 2048; the FFT implementation scales smoothly with width and remains far faster than the naive baseline across the tested range. Finally, Fig. 6 reports the parameter compression ratio as a function of block size for the same MLP, closely tracking the expected approximately 4k reduction relative to a real-valued MLP, with a slight deviation explained by uncompressed bias terms.

These ablations validate the intended efficiency mechanism of EdgeLDR: block size provides a predictable compression knob, and FFT-based computation is the key enabler that converts that structure into practical runtime behavior.

## V. DISCUSSION

The conducted experiments support two complementary takeaways. First, quaternion block-circulant parameterizations can preserve competitive accuracy at substantially reduced parameter budgets by combining (i) cross-channel coupling induced by the Hamilton product with (ii) structured parameter sharing induced by block-circulant operators. In RGB settings, quaternions naturally encode correlated channel groups (e.g., color components). For hyperspectral imaging, quaternion grouping provides a compact mechanism to jointly transform adjacent spectral bands while the block-circulant constraint regularizes channel mixing through cyclic tying. These inductive biases help explain why EdgeLDR can remain competitive even when operating at parameter counts comparable to or smaller than purely real structured baselines.

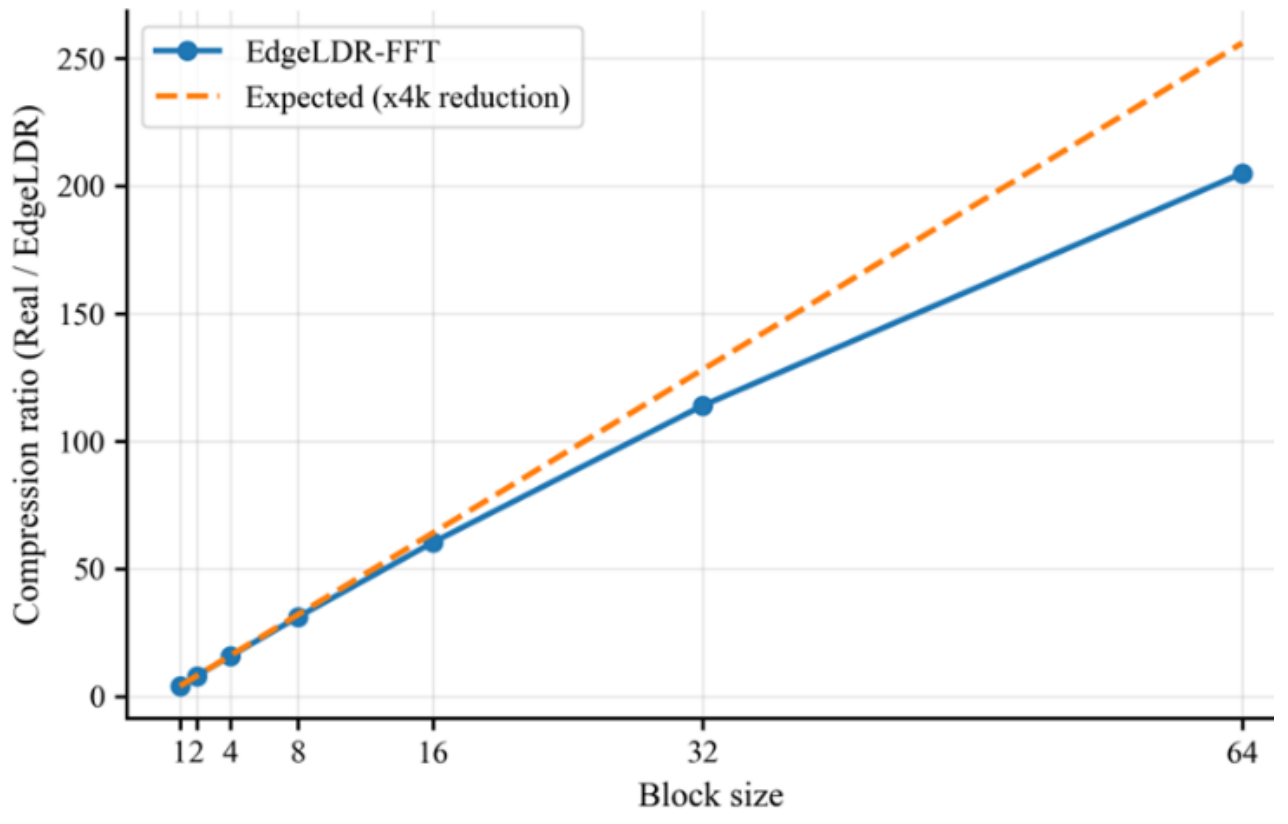

**FIGURE 6. Measured compression ratio vs. block size for EdgeLDR layers (MLP width 1024). Parameter compression (Real / EdgeLDR) as a function of block size, compared against the expected approximately 4k reduction relative to a real-valued MLP (4 real scalars per quaternion). Small deviations from the ideal trend are primarily due to uncompressed bias terms and non-structured components that remain dense.**

Second, the experiments emphasize that parameter compression and algorithmic complexity do not automatically translate into end-to-end latency gains under a straightforward PyTorch implementation. The HSI Transformer results show that, while EdgeLDR layers reduce parameters and can improve OA, their measured GPU/CPU latency can exceed that of dense real baselines. This is consistent with the operational profile of FFT-based layers: they trade dense GEMM throughput for a pipeline that involves batched FFTs, frequency-wise bicomplex arithmetic, and additional memory movement. Without kernel fusion, careful tensor layout, and high-performance FFT primitives, these overheads can dominate for moderate layer sizes.

The ablation study isolates the core computational mechanism: FFT-based quaternion circulant evaluation is essential to make larger block sizes viable. When comparing FFT evaluation against a naive spatial-domain implementation under the same parameterization, FFT maintains a narrow latency band across a wide block-size sweep and yields large empirical speedups. This confirms that the structural constraint is only practically useful when paired with a fast transform implementation.

These observations suggest several practical design guidelines. EdgeLDR layers are most attractive in network regions where (i) channel mixing dominates runtime, (ii) block sizes are large enough to deliver meaningful compression, and (iii) the implementation can amortize FFT overhead through batching and reuse. Conversely, very early layers or narrow classifier heads may benefit more from dense real operators or simpler structured alternatives.

## VI. CONCLUSION

This work presented EdgeLDR, an implementation-oriented framework for quaternion block-circulant fully connected and convolutional layers. By combining quaternion channel coupling with block-circulant structure and using the complex adjoint representation, EdgeLDR layers admit FFT-based evaluation that reduces the cost of the circulant dimension from quadratic to $\mathcal{O}(k \log k)$.

Empirically, controlled benchmarks demonstrate that FFT evaluation is the key enabler: compared to a naive spatial-domain realization, the FFT implementation delivers large speedups and keeps latency stable across a broad block-size sweep. Integrated into compact CNN and Transformer backbones, EdgeLDR layers provide substantial parameter reductions while maintaining competitive accuracy on RGB classification (CIFAR-10/100, SVHN) and hyperspectral classification .

At the same time, the reported CPU/GPU timing results highlight a practical constraint: in a reference PyTorch implementation, end-to-end latency can remain higher than dense real baselines despite fewer parameters, due to FFT overheads and memory-movement costs. Future work will focus on implementation advances - kernel fusion, optimized tensor layouts, and dedicated GPU/edge-accelerator support, to more consistently translate the theoretical efficiency of structured quaternion operators into real deployment gains.